\documentclass[9pt]{article}
\usepackage{spconf,amsmath,graphicx}
\usepackage{mathtools}
\usepackage{amsmath,amssymb,amsfonts}
\usepackage{algorithmic}
\usepackage{xcolor}
\usepackage{dsfont}
\usepackage{sectsty}
\usepackage[nodisplayskipstretch]{setspace}

\title{Directed scattering for knowledge graph-based\\cellular signaling analysis}
\begin{document}
\ninept

\maketitle 
\begin{abstract}

Directed graphs are a natural model for many phenomena, in particular scientific knowledge graphs such as molecular interaction or chemical reaction networks that define cellular signaling relationships. In these situations, source nodes typically have distinct biophysical properties from sinks. Due to their ordered and unidirectional relationships, many such networks also have hierarchical and multiscale structure. However, the majority of methods performing node- and edge-level tasks in machine learning do not take these properties into account, and thus have not been leveraged effectively for scientific tasks such as cellular signaling network inference. We propose a new framework called Directed Scattering Autoencoder (DSAE) which uses a directed version of a geometric scattering transform, combined with the non-linear dimensionality reduction properties of an autoencoder and the geometric properties of the hyperbolic space to learn latent hierarchies. We show this method outperforms numerous others on tasks such as embedding directed graphs and learning cellular signaling networks.

\end{abstract}
\section{Introduction}
\label{sec:intro}


Graph representation learning methods, or methods which learn low-dimensional feature representations for each node in a graph, have shown strong performance and widespread utility for preserving structural information for downstream tasks, including link prediction, community detection, and node classification \cite{Hamilton2017-av}. However, most existing approaches are designed for undirected graphs. This is despite the fact that many relations of interest can be best described as directed; in fact, common benchmarks for graph representation learning, including citation networks, are directed and symmetrized for evaluation. Such real-world relationships also often exhibit hierarchical structure, motivating approaches that leverage hyperbolic geometry \cite{Nickel2017-av, Chami2019-xb, Chami2020-fh}. Notably, many scientific networks are directed but are not regularly benchmarked or represented with direction or latent hierarchies \cite{pratapa_benchmarking_2020}. This includes graphs for cellular signaling, consisting of molecular interactions occurring in a highly ordered series resulting in directed and tree-like relationships \cite{Armingol2021-gn}.

Recently, efforts to incorporate the geometric scattering transform, a generalization of the Euclidean scattering transform to the graph domain, have proven effective in capturing multiscale graph structure and overcoming oversmoothing for undirected graph neural networks \cite{wenkel2022overcoming}. While there exist directed graph neural network-based approaches for supervised tasks \cite{Zhang2021-wq}, we hypothesized geometric scattering would better preserve the graph geometry and improve representations for unsupervised learning. Here, we introduce our framework, Directed Scattering Autoencoder (DSAE), which incorporates a directed version of the geometric scattering transform based on the eigendecomposition of the magnetic Laplacian. Additionally, as the resulting feature representation is high-dimensional, we employ an autoencoder architecture with either Euclidean or hyperbolic geometry to learn a meaningful low-dimensional representation, where the hyperbolic version accounts for additional hierarchical structure. We evaluate DSAE on two previously benchmarked WebKB graphs \cite{Zhang2021-wq} and three cellular signaling graphs \cite{Turei2021-qr, Lo_Surdo2023-ys, Huang2018-da}, comparing against 11 undirected, directed, knowledge graph-based, and hyperbolic-based graph representation learning methods. DSAE performs best for link prediction on four of five graphs due to the ability to account for directedness, higher-order, and hierarchical structure. Second, we describe how to adapt our pipeline by adding additional regularizations to use DSAE for data-driven cellular signaling network inference, demonstrating the efficacy of this approach on a mouse visual cortex dataset with matched spatial data. 

\section{Background}
\label{sec:preliminaries}

\subsection{Magnetic Laplacian}
\label{sec:magnetic_laplacian}
Many methods in (undirected) graph signal processing rely on the eigendecomposition of the graph Laplacian, $L=D-A$. The eigenvalues are viewed as generalized frequencies and the eigenvectors are viewed as generalized Fourier modes. However, it is not straightforward to extend these methods to directed graphs since the naive definition of the graph Laplacian will in general not be symmetric or diagonalizable. While a number of solutions have been proposed \cite{Deri2017-yj, Sardellitti2017-qe, Chung2005-fv}, here, we shall use a complex Hermitian  matrix $L^{(q)}$, known as the magnetic Laplacian (see, e.g, \cite{Furutani2020-lq}). 

To construct the magnetic Laplacian, we let $A$ be the (asymmetric) adjacency matrix of a directed graph $G = (V,E)$, $N =|V|$, let $A^{(\text{sym})} = \frac{1}{2}(A + A^{T})$ be its symmetrized counterpart, and let $D^{(\text{sym})}$ be the diagonal degree matrix corresponding to $A^{(\text{sym})}$, i.e. $D^{(\text{sym})}_{j,j} = \sum_{k=0}^{N-1} A^{(\text{sym})}_{j,k}$, and $D^{(\text{sym})}_{j,k} = 0$ if $j \neq k$. Then, we capture directional information via the phase matrix $\Theta^{(q)}$, where for $q\geq 0$, we define $\Theta^{(q)} = 2\pi q(A - A^{T}).$ Letting $i=\sqrt{-1}$, this allows us to define the complex Hermitian adjacency matrix by $$H^{(q)} = A^{(\text{sym})} \odot \exp(i\Theta^{(q)}),$$ where $\odot$ denotes the Hadamard product and the exponentation is performed term-by-term. We then define the unnormalized and 
normalized magnetic Laplacians by $$L_U^{(q)} = D^{(\text{sym})} - H^{(q)}\quad\text{and}\quad L_N^{(q)} =  {D^{(\text{sym})}}^{-1/2}L_U^{(q)}{D^{(\text{sym})}}^{-1/2}
$$ respectively. Since $\Theta^{(q)}$ is skew-symmetric, it is clear that both $L_U^{(q)}$ and $L_N^{(q)}$ are Hermitian, and Theorem 1 of \cite{Zhang2021-wq} shows that they are positive semi-definite. Therefore, they both admit orthonormal bases of eigenvectors with non-negative eigenvalues $\mathbf{u}_k$, $0\leq k\leq N-1$, $L^{(q)}\mathbf{u}_k=\lambda_k\mathbf{u}_k$, $0\leq \lambda_0\leq \lambda_1\leq\ldots\leq \lambda_{N-1}$ (where $L^{(q)}$ is either  $L^{(q)}_U$ or $L^{(q)}_N$).

\subsection{The Geometric Scattering Transform}
\label{sec:geometric_scattering}
The geometric scattering transform  \cite{Gao2018-nz,gama2018diffusion,zou2020graph} is a recently proposed multiscale, multi-order signal transform that applies an alternating sequence of wavelet transforms and non-linear activations to an input signal $\mathbf{x}:V\rightarrow \mathbb{R}$, building off of analogous work for functions defined on $\mathbb{R}^n$ \cite{Mallat2011-kh}. 
Given a wavelet frame, $\mathcal{W}_J = \{W_j\}^J_{j=0} \cup \{A_J\}$ (where $W_j$ and $A_J$ are $n\times n$ matrices) and an input signal $\mathbf{x}:V\rightarrow\mathbb{R},$ which we identify with the vector $\mathbf{x}[k]=\mathbf{x}(v_k)$, we define first- and second-order scattering coefficients of $\mathbf{x}$, for $0\leq j_1\leq j_2\leq J$, by 
$$
S[j_1]\mathbf{x}=A_JM W_{1}j\mathbf{x}
,\quad S[j_1,j_2]\mathbf{x}=A_JM W_{j_2}M W_{j_1}\mathbf{x},
$$  where $M$ denotes the entry-wise modulus (absolute value). The zeroth-order coefficient is defined simply by 
$A_J\mathbf{x}$. 

\subsection{Hyperbolic geometry}
\label{sec:hyperbolic_geometry}
Hyperbolic geometry is a non-Euclidean geometry that embeds tree-structured data with arbitrarily low distortion \cite{Sarkar2012-av}. Hyperbolic spaces have constant negative curvature, where curvature measures how a geometric object deviates from a flat plane. Here, we use the $d$-dimensional Poincaré ball model with constant negative curvature $-c\ (c>0)$, defining smooth manifold ($\mathds{D}^{d,c}, g_x)$, where $\mathds{D}^{d,c} = \{x \in \mathds{R}^d : \|x\|^2 < \frac{1}{c}\}$ and $g_x$ is the Poincaré metric tensor \cite{Nickel2017-av}. 
The exponential map $\textrm{exp}_x^c: \mathcal{T}_{x}^{c} \mapsto \mathds{D}^{d,c}$ and the logarithmic map $\textrm{log}_x^c: \mathds{D}^{d,c} \mapsto \mathcal{T}_{x}^{c}$ map between the tangent space at $x$ ($\mathcal{T}_{x}^{c}$) and $\mathds{D}^{d,c}$ \cite{Ganea2018-eb}. The closed-form for these maps at the origin are:
$$
\textrm{exp}_0^c(v)\hspace{-.017in}=\hspace{-.016in}\textrm{tanh}(\sqrt{c}\|v\|)\frac{v}{\sqrt{c}\|v\|},
\textrm{log}_0^c(y)\hspace{-.017in}=\hspace{-.016in}\textrm{tanh}^{-1}(\sqrt{c}\|y\|)\frac{v}{\sqrt{y}\|y\|}.
$$
\subsection{Related Work}
\label{sec:relatedwork}

Learning meaningful representation of nodes for directed graphs has relevant applications in works related to directed graph neural networks \cite{Zhang2021-wq} and knowledge graph embeddings \cite{Bordes2013-av}, where knowledge graphs are defined as directed, heterogeneous multigraphs. Additionally, \cite{Chew2022-or} recently explored a directed-graph version of the graph scattering transform on synthetic data. Concurrently, there have been developments in hyperbolic geometry that show potential for preserving latent hierarchies \cite{Nickel2017-av, Ganea2018-eb, Chami2019-xb}. While some preliminary work incorporates hyperbolic geometry into directed graph learning \cite{Chami2020-fh}, no work has accounted for this geometry for biological networks. Most embedding approaches do not evaluate on biological networks, and benchmarks for these networks often test on undirected (or symmetrized) versions \cite{pratapa_benchmarking_2020}. This is despite the fact that much biological analysis considers the directionality of the relationship (e.g., between two interacting molecules \cite{Armingol2021-gn, Efremova2020-df}).



\section{Methods}
\begin{figure}[htb]

\begin{minipage}[b]{1.0\linewidth}
  \centering
  \centerline{\includegraphics[width=86mm]{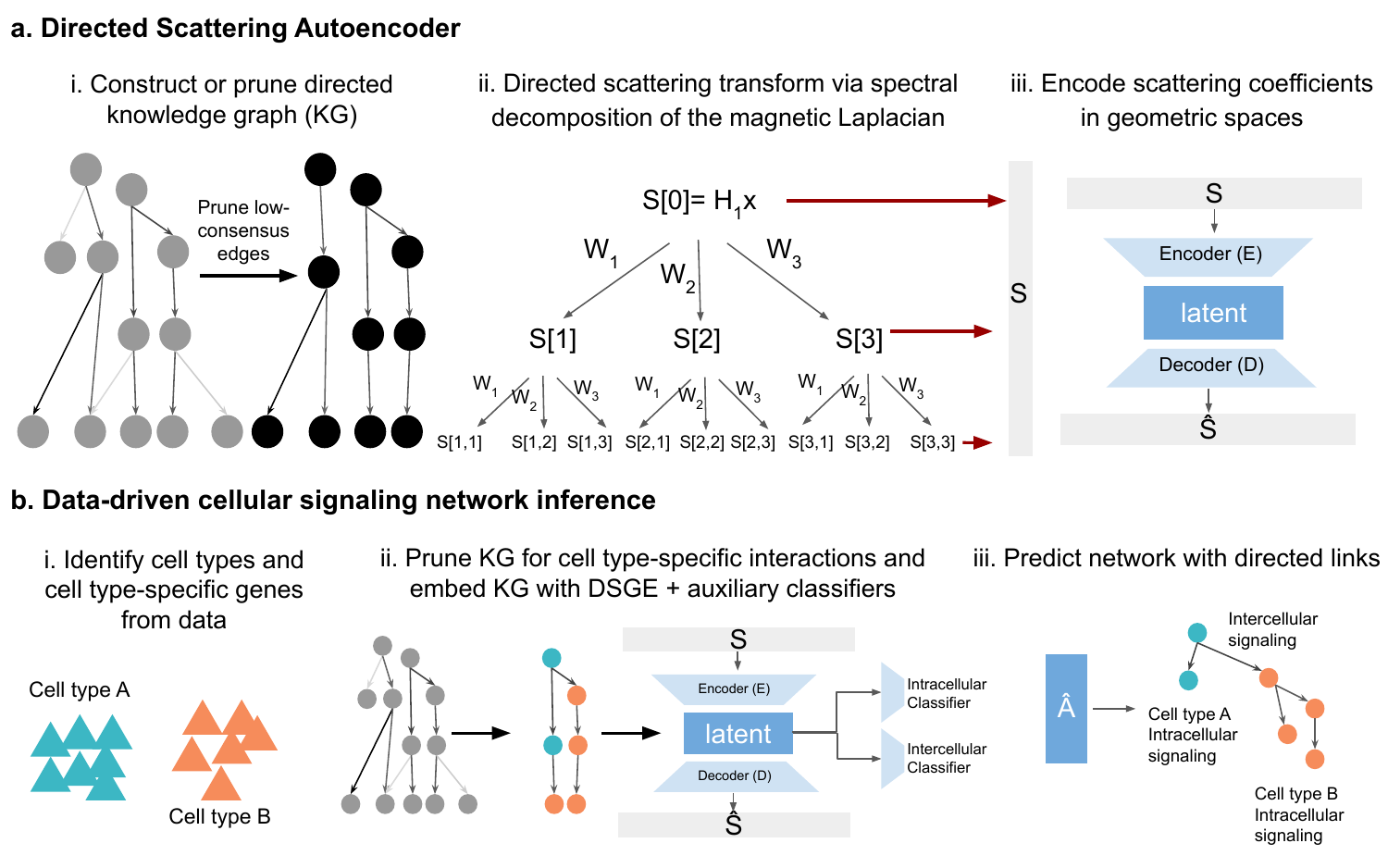}}
\end{minipage}
\caption{DSAE and cellular signaling network inference pipeline.}
\label{fig:2_method}
\end{figure}

Our goal is to learn a representation of the nodes on a directed graph that preserves both directional and hierarchical information (Fig. \ref{fig:2_method}a).
To accomplish this, we introduce a method which first applies a directed-graph version of the scattering transform, then applies an autoencoder restructure the scattering coefficients into a compressed representation for improved performance on downstream tasks.

First, following the lead of \cite{Chew2022-or}, we use the (normalized or unnormalized) magnetic Laplacian $L^{(q)}$ to construct a directed-graph wavelet frame $\mathcal{W}_J = \{W_j\}^J_{j=0} \cup \{A_J\}$. As in Section \ref{sec:magnetic_laplacian}, the $q$ parameter associated to the magnetic Laplacian will determine the way in which $L^{(q)}$ encodes directional information. As in traditional wavelet constructions, the $j$ parameter denotes the scale of the wavelet $W_j$ with $J$ being the maximal scale.

More specifically, for $t\geq 0$, we define the graph heat-kernel by  $H_t=\sum_{k=0}^{N-1}e^{-\lambda_kt}\mathbf{u}_k\mathbf{u}_k^*$, 
 where $^\star$ denotes conjugate transpose, and define a wavelet frame for a fixed $J\geq 0$, by $
 \mathcal{W}_J = \{W_j\}^J_{j=0} \cup \{A_J\}$
 where $W_0=\text{Id}-H_1,$ $A_J=H_{2^J}$, and  $W_j=H_{2^{j-1}}-H_{2^j}$ for $1\leq j\leq J$.  

We will assume that we are given $C$ signals $\mathbf{x}_1,\ldots,\mathbf{x}_C$ and will compute the zeroth-, first-, and second-order scattering coefficients of each $\mathbf{x}_i$ using the formulas from Section \ref{sec:geometric_scattering}. However, inspired by \cite{wenkel2022overcoming}, we use $H_1$ instead of $A_J$ in the final filtering (e.g., we compute second-order scattering coefficients by $H_1M W_{j_2}M W_{j_1}\mathbf{x}$). We then let $S[v]$ denote the concatenation of all zeroth-, first-, and second-order scattering coefficients  evaluated at the vertex $v$ and let $S(G)=\{S[v]: v\in V\}$.

The node representations $S[v]$ are typically redundant and unnecessarily high-dimensional. Therefore, in order to reduce the dimension of our preliminary scattering representation, we apply an autoencoder $D\circ E$ so that 
$$S(G)\approx\hat{S}(G)= D(E(S(G)).$$ In particular, while training our autoencoder, we aim to minimize the mean squared error (MSE) reconstruction loss 
$$
\mathcal{L}_{\text{recon}}=\|S(G) - \hat{S}(G) \|^2_2=\sum_{v\in V}\|S[v]-\hat{S}[v]\|_2^2.$$  
 We consider two versions of the feed-forward encoder and decoder layers to learn a $d$-dimensional embedding, where $d\ll$ the dimensionality of $S[v]$. First, we consider matrix operations carried out via Euclidean geometry (DSAE-Euc). Second, since each $S[v]$ has inherent tree-like structure (Fig. \ref{fig:2_method}), we consider an analog to Euclidean operations for hyperbolic space to preserve hierarchical structure: M\"obius addition $\oplus_c$ and M\"obius matrix-vector multiplication $\otimes_c$ \cite{Ganea2018-eb} (DSAE-Hyp) . For the encoder in the latter case, we map $S[v]$ to $\mathds{D}^{d,c}$, the $d$-dimensional Poincaré ball of radius $c$, via the exponential map $h_0^v = \textrm{exp}_0^c(S[v])$ (see \cite{Ganea2018-eb, Chami2019-xb}), then employ hyperbolic feed-forward layers, i.e.
$$
h_\ell^v \leftarrow \sigma((W_\ell \otimes_c h_{\ell-1}^v) \oplus_c \textrm{exp}_0^c(b_\ell))
$$
for hidden representation $h_\ell$ at layer $\ell$, 1 $\leq \ell \leq L$, where $L$ is the number of layers. $W_\ell$ is the weight matrix (which is shared across $v$), $b_\ell$ is the bias, and $\sigma$ is the activation. For the decoder, we use the same construction and map to the Euclidean space for reconstruction via $\textrm{log}^c_0(h_L^v)$. We view the encoder output $E(S[v])$ as the hidden representation of each vertex $v$. In Section \ref{sec: Exp Lp}, we show this representation may be used for tasks such as link prediction. 


When applying our method, we may also use an additional network to regularize our autoencoder. When doing so, we assume that there is some property of interest $p(v)$ which is important for downstream tasks. Therefore, we train an additional network $F$ (or possibly multiple additional networks) so that $F(E(S[v]))\approx p(v)$ and penalize this network by 
$$
\mathcal{L}_{\text{prop}}=\sum_{v\in V} \text{dist}(p(v),F(E(S[v]))),
$$
where the distance function will vary depending on the task of interest. This regularizing network helps our autoencoder learn a compressed representation of the scattering coefficients which preserves information relevant for a given application but discards less relevant information. In Section \ref{sec: regularization example}, we will provide an example of how such regularizing networks may be applied to cellular signaling networks. 

\section{Link Prediction in Directed Graphs}\label{sec: Exp Lp}
We evaluate the performance of our method on the task of link direction prediction on two WebKB graphs modeling links between websites at different universities \cite{Zhang2021-wq}, and three cellular signaling graphs \cite{Turei2021-qr, Lo_Surdo2023-ys, Huang2018-da}. In the latter case, the nodes are genes and the edges encode signaling relationships, including transcriptional regulation (within cells), or protein-protein binding after secretion (between cells). While constructing the biological graphs, we prune the network for interaction confidence, restrict the graph to the largest connected component, and give all remaining edges unit weight so as not to overemphasize well-studied genes. 
 
We note that all graphs are directed and hierarchical; less than 20\% of edges are bidirectional for the website graphs, and less than 5\% of the edges are bidirectional for the biological graphs. Further, the Krachkardt hierarchy score (Khs) \cite{Krackhardt1994-av} is over 0.75 (between 0 and 1, where 1 is most hierarchical and 0 least hierarchical) for all graphs, and the Ollivier-Ricci curvature \cite{Ollivier2009-av} (averaged for all edges) is negative. The $k$-NN graphs of the directed scattering coefficients also have negative average Ollivier-Ricci curvature (data not shown). Further details on the dataset are available in Table \ref{table:1_summary}.  

\setlength{\tabcolsep}{1pt}
\begin{table}[!ht]
\begin{center}
\begin{tabular}{||c | c c c c c c||} 
\hline
& Type & \# Nodes & \# Edges & Rec. & Khs &\ ORC \\
\hline\hline
Texas & WebKB & 183 & 325 & 18\% & 0.91 &\ -0.14\\ [0.5ex]
Cornell & WebKB & 183 & 298 & 12\% & 0.96 &\ -0.30\\[0.5ex]
OmniPath &
\begin{tabular}{@{}c@{}}Post-translational\\and  transcriptional\end{tabular}
& 8909 & 58919 & 4\% & 0.76 &\ -0.51\\[0.5ex]
SIGNOR\_ppi & Post-translational & 4743 & 11109 & 2\% & 0.92 &\ -0.56\\[0.5ex]
iPTMnet & Post-translational & 1642 & 3650 & 0\% & 0.98 &\ -0.44\\[0.5ex]
\hline
\end{tabular}
\end{center}
\caption{Statistics for each directed network. Rec: Reciprocity, Khs: Krackhardt hierarchy score. ORC: Ollivier-Ricci curvature.}
\label{table:1_summary}
\end{table}

On each of these graphs, we are given a partially complete version of the network (i.e., the original graph, but with some edges removed) and the task is to predict whether a given edge, $(u,v)$ or its reciprocal edge $(v,u)$ belongs to the original network. As baseline methods, we considered shallow approaches -- node2vec \cite{Grover2016-gm} and Poincaré map (PM) \cite{Nickel2017-av}; graph neural network approaches -- graph autoencoder (GAE) \cite{Kipf2016-xy},  hyperbolic graph convolutional network (HGCN) \cite{Chami2019-xb}, and directed graph neural network MagNet \cite{Zhang2021-wq}; a  knowledge graph-based approach -- TransE \cite{Bordes2013-av}; and the undirected graph scattering transform (UDS) \cite{Gao2018-nz}. When applicable, we use a single input signal chosen to be a standard Gaussian random vector.

\setlength{\tabcolsep}{4pt}
\begin{table}[!ht]
\begin{center}
\begin{tabular}{||c | c c c c c ||} 
 \hline
Method & Texas & Cornell & OmniPath & SIGNOR & iPTMnet \\
\hline\hline
DSAE-Hyp & \textbf{0.915} & \textbf{0.936} & \underline{0.846} & \textbf{0.905} & 0.983 \\
DSAE-Euc &\underline{0.913} & 0.809 & \textbf{0.870} & 0.893 & \underline{0.988} \\
node2vec & 0.860 & 0.714 & 0.608 & 0.703 & 0.804 \\
PM & 0.591 & 0.574 & 0.545 & 0.579 & 0.530 \\
PM-D & 0.617 & 0.678 & 0.581 & 0.564 & 0.602 \\
GAE & 0.807 & 0.584 & 0.716 & 0.780 & 0.873 \\
GAE-D & 0.806 & 0.695 & 0.712 & 0.793 & 0.881 \\
HGCN & 0.729 & 0.639 & 0.586 & 0.711 & 0.816 \\
HGCN-D & 0.799 & 0.754 & 0.714 & 0.733 & 0.910 \\
MagNet & 0.888 & 0.930 & 0.831 & \underline{0.898} & \textbf{0.989} \\
TransE & nan & nan & 0.793 & nan & nan \\
TransE-E & 0.801 & \underline{0.933} & 0.777 & \underline{0.898} & \underline{0.988} \\
UDS-AE & 0.845 & 0.893 & 0.668 & 0.783 & 0.976 \\
 \hline
\end{tabular}
\end{center}
\caption{Link Direction Prediction mean AUROC. -D refers to implementation with asymmetric matrices -E refers to implementation with edge attributes. nan refers to undefined runs (no edge attributes). Best performance bolded, second best underlined.\vspace{-0.4cm}}
\label{table:2_results}
\end{table}
Table \ref{table:2_results} summarizes comparisons of DSAE-Hyp and DSAE-Euc to the other approaches. Both versions of DSAE perform quite well. DSAE-Hyp is the top performing method on Texas, Cornell, and SIGNOR and is the second best on OmniPath. DSAE-Euc is the top method on OmniPath and second best on Texas and iPTMnet.


\subsection{Ablation experiment}
To test the importance of the autoencoder and the robustness to $q$ (the charge parameter) and $J$ (the scale parameter), we performed a series of ablations (shown in Table \ref{table:3_ablation}) with DSAE-Hyp, which performed best on average. Ablations showed q=0.0 (which does not take into account the edge direction) performed worse than $q=0.1$ and $q=0.2$ at all scales (with all other optimal hyperparameters), showing that edge direction was useful for link prediction. Directed scattering without the autoencoder (no AE) performed worse than the full DSAE method on average (setting $J$ and $q$ to be the same as the optimal performing version of DSAE). 

\begin{table}[!ht]
\begin{center}
\begin{tabular}{||c | c c c c c ||} 
 \hline
 Method & Texas & Cornell & OmniPath & SIGNOR & iPTMnet\\ [0.5ex] 
 \hline\hline
no AE & 0.903 & 0.929 & 0.843 & 0.903 & 0.981\\
q=0.0 J=5 & 0.750 & 0.641 & 0.519 & 0.549 & 0.922 \\
q=0.0 J=10 & 0.747 & 0.642 & 0.518 & 0.561 & 0.921 \\
q=0.0 J=15 & 0.751 & 0.641 & 0.518 & 0.564 & 0.921 \\
q=0.1 J=5 & 0.906 & 0.930 & \textbf{0.846} & \textbf{0.905} & 0.980 \\
q=0.1 J=10 & \textbf{0.915} & 0.932 & \textbf{0.846} & \textbf{0.905} & 0.980 \\
q=0.1 J=15 & 0.911 & 0.932 & 0.845 & 0.904 & 0.979 \\
q=0.2 J=5 & 0.900 & 0.935 & 0.833 & 0.904 & \textbf{0.983} \\
q=0.2 J=10 & 0.899 & \textbf{0.936} & \textbf{0.846} & 0.904 & 0.982 \\
q=0.2 J=15 & 0.899 & 0.935 & 0.845 & 0.903 & 0.982 \\
 \hline
\end{tabular}
\end{center}
\caption{Ablation Link Direction Prediction mean AUROC. no AE refers to scattering without autoencoder. Best performance bolded.}
\label{table:3_ablation}
\end{table}
\vspace{-0.4cm}

\subsection{Training details}
We trained with an Adam / RiemmanianAdam optimizer for 50 epochs with a patience of 10 epochs with latent dimension $d=128$. The train/val/test split was 85/5/10, where the training subgraph was first built from a minimum spanning tree with edges added until split proportion met. We optimized for the following: walk length and number of walks (node2vec); learning rate, bias, dropout, number of layers, activation, and weight decay (remaining methods); attention versus no attention (GAE); q (MagNet, DSAE-Hyp, DSAE-Euc); c (HGCN, Poincaré map, DSAE-Hyp); and J (DSAE-Hyp, DSAE-Euc). Results are reported as mean AUROC of five runs\footnote{github.com/KrishnaswamyLab/Directed-Hierarchical-Gene-Networks}. 

\section{Inferring a directed graph from  data}\label{sec: regularization example}
Here, we show that we can use DSAE to infer active prior knowledge signaling and discover \textit{de novo} signaling connections in a data-driven manner (Fig. \ref{fig:2_method}b). Given single-cell RNA-sequencing data, or high-dimensional point-cloud data where the observations are cells and the features are gene expression measurements, we want to understand what genes are signaling to each other within and between cell types based on their expression and prior knowledge. Specifically, we aim to infer a directed gene-gene graph defined by two within-cell type gene networks connected by between-cell type interactions. While most prior work identifies correlated or mutually informative gene pairs \cite{Armingol2021-gn, Hu2021-yj}, this is, to our knowledge, the first approach to infer the entire directed gene network between cell types.

\subsection{Constructing the cellular signaling network}
We let $G=(V,E)$ be a directed graph where each vertex $v$ represents a gene and edges represent genes which interact within or between cells (such as OmniPath or SIGNOR). We assume each gene corresponds to a certain cell type and let $V_i$ denote the set of all genes corresponding to cell type $i$.
Given two cell types $i$ and $j$, 
we let $G_{i,j}=(V_{i,j},E_{i,j})$ be the largest connected component of the induced subgraph with vertices $V_i\cup V_j$. 

\subsection{Applying DSAE to Cellular Signaling Networks}
In order to structure the latent space for cellular signaling inference,
when applying DSAE to the graph $G_{i,j}$ constructed above,  we regularize our autoencoder with two additional classifiers that predict cell labels from hidden representation $E(S[v])$. In particular, we have an intracellular classifier, which predicts the cell type, and an intercellular classifier, which predicts whether or not the node is capable of intercellular communication. Each of these classifiers are penalized by supervised contrastive losses $\mathcal{L}_{\text{intracellular}}$ and $\mathcal{L}_{\text{intercellular}}$ \cite{Khosla2020-ic}. The intracellular classifier ensures that nodes from the same cell type remain close. The intercellular classifier, on the other hand, ensures that nodes which are capable of intercellular communication and are connected to nodes from the opposite cell type remain close. These additional classifiers allow us to jointly learn a shared representation for both unsupervised reconstruction and supervised intracellular and intercellular classification. 
The total loss is then 
\begin{equation}\label{eqn: loss}
\mathcal{L}_{\text{total}} = \alpha \mathcal{L}_{\text{recon}} + \beta \mathcal{L}_{\text{intracellular}} + \gamma \mathcal{L}_{\text{intercellular}}.\end{equation}

After training DSAE, we then replace the initial $E_{i,j}$ with a new edge set. First, we add an undirected edge between vertices $v_i$ and $v_j$ if $v_i$ is amongst the $k$-th nearest (in the embedded space) neighbors of $v_j$ or vice-versa. Then, we assign a direction 
via a linear (logistic regression) classifier trained on the edges from the original $E_{i,j}.$

\subsection{Case study: Mouse visual cortex scRNA-seq data}
\begin{figure}[htb]

\begin{minipage}[b]{1.0\linewidth}
  \centering
  \centerline{\includegraphics[width=86mm]{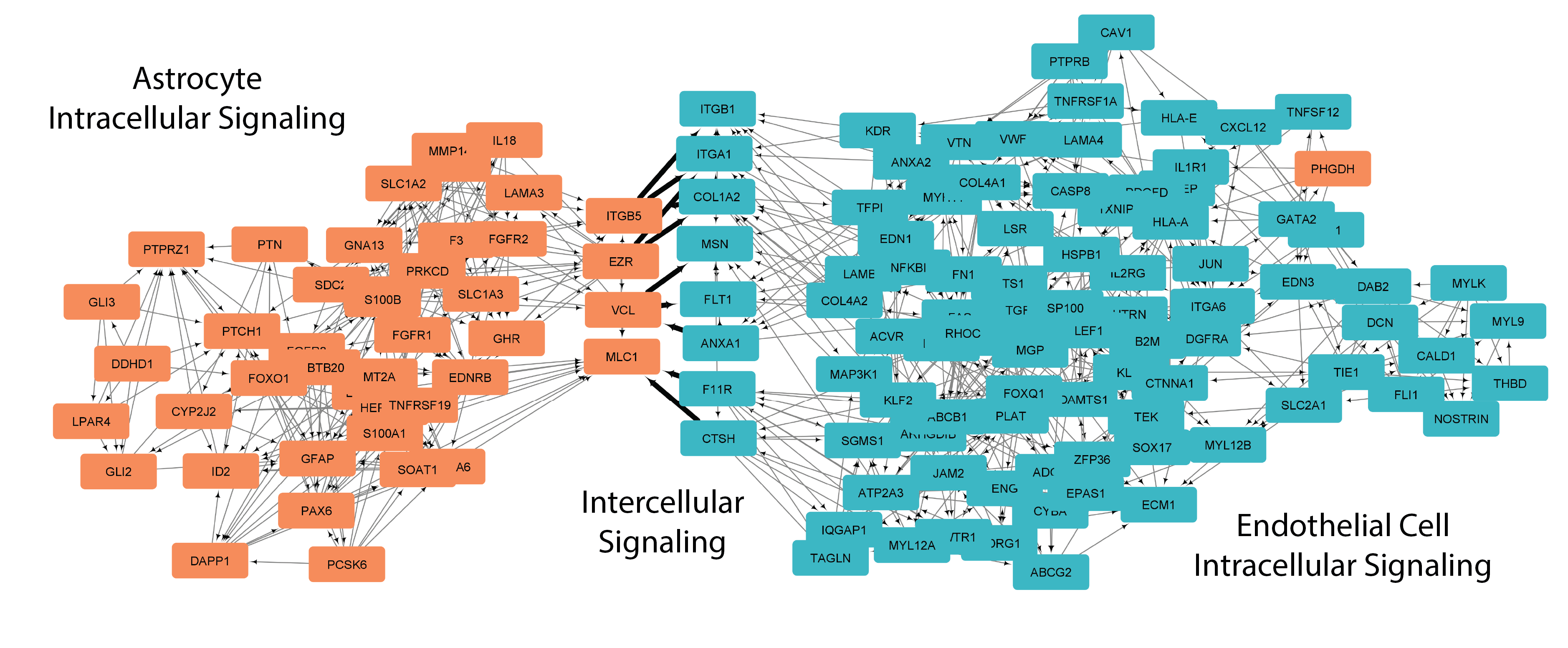}}
\end{minipage}
\caption{Learned cellular signaling network between cell types.}
\label{fig:5_network}
\end{figure}

We take two previously annotated cell types (astrocytes and endothelial cells) from an scRNA-seq dataset of the mouse visual cortex \cite{Tasic2016-av} and identify cell type-specific genes as genes with log2 fold change $>$ 2 and p-value $<$ 0.05 in a given cell type from a Wilcoxon rank sum test. We prune the OmniPath network (version 1.0.5) to only these genes and label nodes based on cell type and intercellular connection, then apply DSAE-Euc with $(\alpha,\beta,\gamma)=(10,5,1)$ in \eqref{eqn: loss}, resulting in an embedding informed by intracellular and intercellular connections in addition the original graph $G$. We then build a directed network with a $k$-NN graph ($k$=5) and visualize the network with Cytoscape \cite{Shannon2003-cv} (Fig. \ref{fig:5_network}). We see that four astrocyte genes and eight endothelial cell genes are involved in intercellular signaling.

\begin{figure}[htb]

\begin{minipage}[b]{1.0\linewidth}
  \centering
  \centerline{\includegraphics[width=86mm]{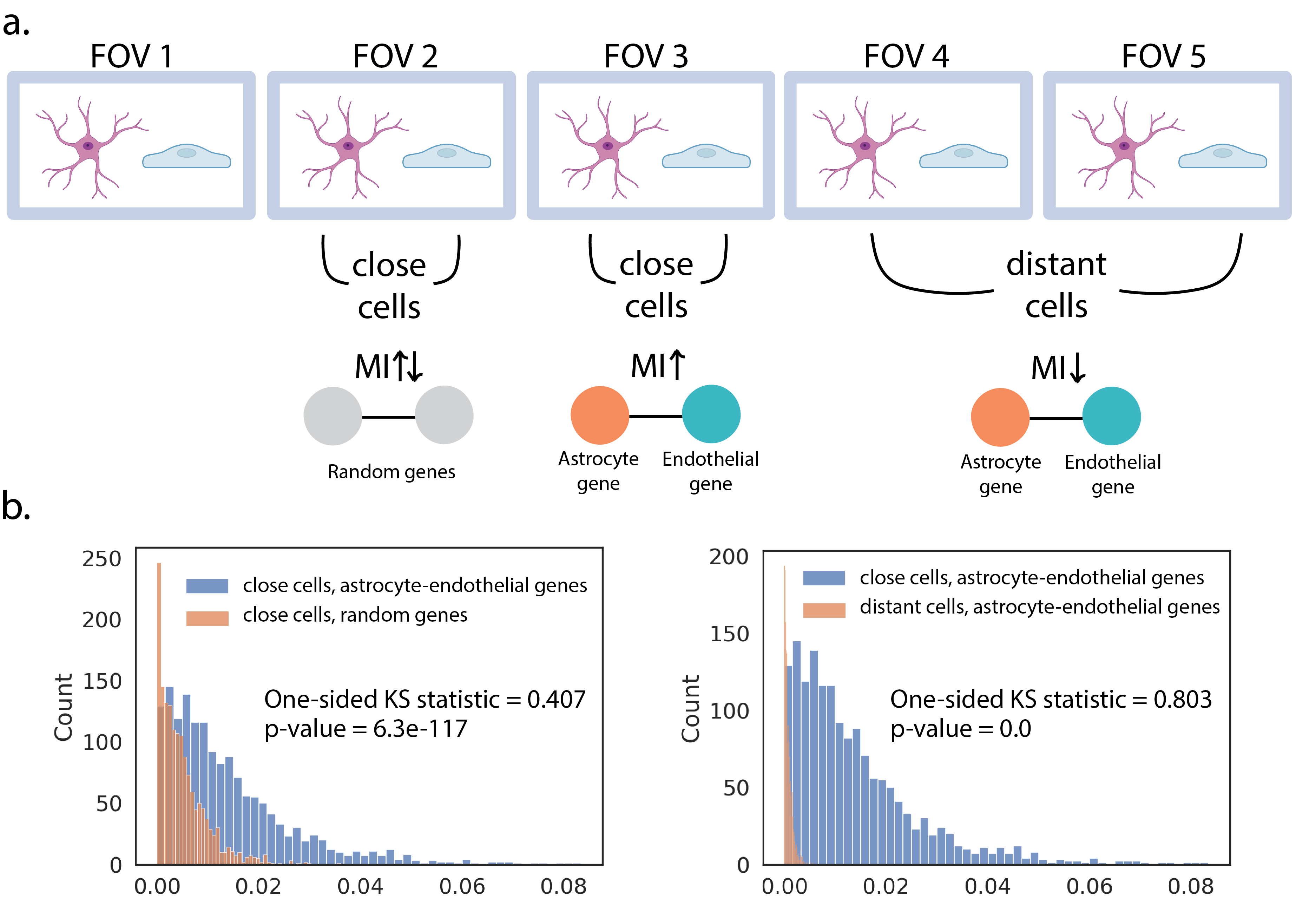}}
\end{minipage}
\caption{Evaluation of learned network with spatial data.}
\label{fig:6_spatial_test}
\end{figure}

Following \cite{Hu2021-yj}, we validate our network with matched mouse visual cortex data with spatial resolution \cite{Eng2019-av}. We reasoned cells in the same spatial field-of-view (FOV) are close and are more likely to signal to each other than cells in different FOVs. True signaling pathway genes should have higher mutual information (MI) of spatial expression in close cells than distant cells, whereas random gene pairs may have high or low MI between close cells (Fig. \ref{fig:6_spatial_test}a). We confirm that genes in our network have a higher MI than random genes between close astrocytes and endothelial cells (one-sided Kolmogorov-Smirnov (KS) test \cite{Massey1951-st} statistic = 0.407, p=6.3e-117). Genes in our network also have a higher MI between close astrocytes and endothelial cells than between distant astrocytes and endothelial cells (one-sided KS test statistic = 0.803, p=0.0) (Fig. \ref{fig:6_spatial_test}b).

\section{Conclusion}
We have introduced DSAE, a novel method for node embeddings which trains a (regularized) autoencoder on top of a directed-graph variation of the scattering transform. The autoencoder learns a compressed representation and generates improved embeddings, in particular with the hyperbolic variation that better represents latent tree-like motifs common to real-world graphs such as biological networks. We show that DSAE is effective at representing five directed graphs due to the ability to preserve directed, multiscale, and hierarchical graph structure. Moreover, when properly regularized, DSAE preserves relevant information (for a given task), including learning active and novel gene-gene interactions for cellular signaling analysis. We emphasize this is the first known application of directed scattering to a real-world setting, providing a key contribution to both graph signal processing and biomedical research.




\newpage
\bibliographystyle{IEEEbib}
\bibliography{references}

\end{document}